\documentclass{bmvc2k}
\usepackage{pdfpages}

\title{Amodal segmentation just like doing a jigsaw}

\addauthor{Xunli Zeng}{}{1}
\addauthor{Jianqin Yin}{}{2}

\addinstitution{
}
\addinstitution{
}

\runninghead{Running Head}{Amodal Segmentation Just Like Doing A Jigsaw}

\begin{document}
	
	\maketitle
	
	\begin{abstract}
		Amodal segmentation is a new direction of instance segmentation while considering the segmentation of the visible and occluded parts of the instance. The existing state-of-the-art method uses multi-task branches to predict the amodal part and the visible part separately and subtract the visible part from the amodal part to obtain the occluded part. However, the amodal part contains visible information. Therefore, the separated prediction method will generate duplicate information. Different from this method, we propose a method of amodal segmentation based on the idea of the jigsaw. The method uses multi-task branches to predict the two naturally decoupled parts of visible and occluded, which is like getting two matching jigsaw pieces. Then put the two jigsaw pieces together to get the amodal part. This makes each branch focus on the modeling of the object. And we believe that there are certain rules in the occlusion relationship in the real world. This is a kind of occlusion context information. This jigsaw method can better model the occlusion relationship and use the occlusion context information, which is important for amodal segmentation. Experiments on two widely used amodally annotated datasets prove that our method exceeds existing state-of-the-art methods. The source code of this work will be made public soon.
	\end{abstract}
	
	\section{Introduction}
	\label{sec:intro}
\begin{figure*}
		\begin{center}
			\includegraphics[width=12cm]{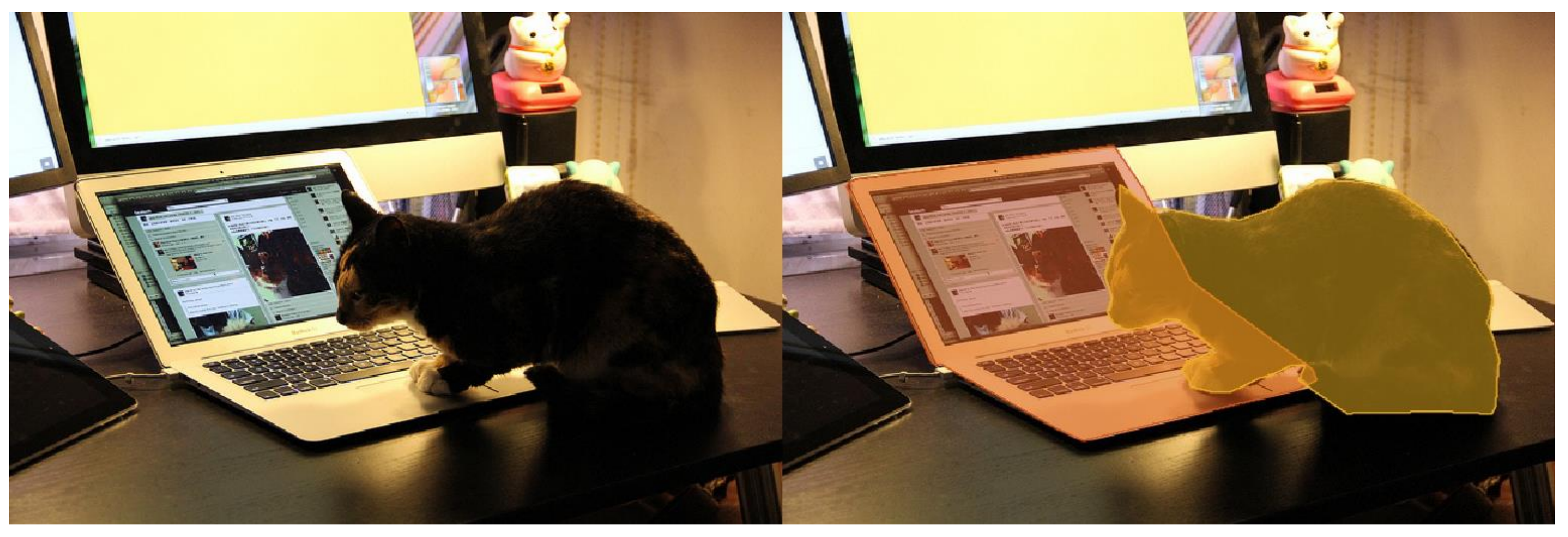}
		\end{center}
		\caption{Humans have powerful amodal perception capabilities, just like when seeing the occluded scene in the picture on the left: a cat lying prone in front of the laptop. You can still perceive the complete shape of the laptop, as shown on the right.}
		\label{fig1}
	\end{figure*}

	When you are walking on the street and about to turn at an intersection, you see a bicycle wheel suddenly appearing in front of you, and you know that there is a cyclist behind the wall at the moment, although you don’t see him. Then you stay in place, waiting for the cyclist to pass first. People often have such scenes in their lives. But this is particularly difficult for robots. Because people have a powerful visual system, they can perceive the overall target object only through some local areas of the target object. In order for the robot to also have the overall visual ability to perceive the object through the local, visible information of the object (show as Fig. \ref{fig1}), the task of amodal segmentation~\cite{1li2016amodal} was proposed.
	
	Amodal segmentation is a complex high-level perception task. It needs to segment both the visible part of the target object and the occluded part of the target object. The amodal mask can be considered to be composed of the visible mask and occlusion mask of the instance object. From the perspective of amodal segmentation task classification, the current research results can be roughly divided into two categories. The first category thinks that amodal segmentation is a single task. These models obtain amodal perception ability by learning the amodal mask that people have annotated on the dataset and directly infers the target’s amodal mask (ASN~\cite{3qi2019amodal}, SLN~\cite{5zhang2019learning}). Mask R-CNN~\cite{6he2017mask} often is trained with amodally annotated dataset as the baseline. ASN~\cite{3qi2019amodal} adds whether there is an occlusion in the branch prediction area and uses the judgment information of whether there is occlusion to assist amodal segmentation. SLN~\cite{5zhang2019learning} uses a new representation of a semantics-aware distance map instead of the mask as the prediction target to segment the amodal mask of the instance. The second category divides amodal segmentation into two parts to complete. For example, VRS\&SP~\cite{9yuting2021amodal} first segment the visible part of the target object and then add shape prior information to infer the amodal mask. ORCNN~\cite{4follmann2019learning} predicts the amodal part and the visible part separately and subtracts the two to get the occluded part.
	
	Amodal segmentation is a complex task. Dividing it into two parts will help reduce the granularity of the model and improve the prediction effect of each part. In this way, the second category of methods is dominant. When the amodal segmentation is divided into two parts to complete, how to decompose is the key to determining the effect of the model. There is no intersection between the visible mask and the occlusion mask of the same instance, so the predictions of the two parts in the multi-task branch are decoupled, which will make each branch focus on the modeling of the object. What's more, we believe that there are certain rules in the occlusion relationship in the real world. For example, in the occlusion relationship formed by the dinner plate and bread, it is often that the bread obscures the dinner plate. This is a kind of occlusion context information. The exploration of this context can help the amodal segmentation. These motivate us to propose a multi-task branch and combined it with the occlusion relationship modeling amodal segmentation method. The method firstly uses multi-task branches to first obtain two pieces of the instance (visible mask and occlusion mask). And then model the occlusion relationship. Finally, utilize the modeled occlusion relationship and stitch the two parts to get the complete jigsaw of the instance (amodal mask).
	
	Our paper's contribution could be summarized as the following aspects:
	\begin{itemize}
		\item We propose to use multi-task branches to first obtain two pieces of the instance (visible part and occlusion  part). Then stitch the two parts to get the complete jigsaw of the instance. This makes each branch focus on the modeling of the object.
		\item In order to use the occlusion context information to help amodal segmentation, we modeled the occlusion relationship. And we use the modeled occlusion relationship when completing the jigsaw of the instance.
		\item The experimental results on two widely used datasets (KINS and COCOA cls) show that our method exceeds the current state-of-the-art methods.
	\end{itemize}
	\section{Related Work}
	\subsection{Instance Segmentation}
	As one of the four basic tasks of computer vision (classification, object detection, semantic segmentation, and instance segmentation), predecessors have done a lot of research. Among these works~\cite{17hariharan2014simultaneous,18dai2016instance,19li2017fully,20chen2018masklab,21pinheiro2016learning}, the most representative one is Mask R-CNN ~\cite{6he2017mask} based on the Faster R-CNN \cite{16ren2016faster} object detection framework, which sends the features extracted by the Backbone into The RPN generates proposals and uses RoIAlign feature pooling to obtain fixed-sized features of each proposal. Because of the fixed-sized features, the accuracy of segmentation is improved. PANet~\cite{11liu2018path} makes the information path between the bottom-up and the top-level features of the deep network shorter by using bottom-up path augmentation. Mask scoring RCNN~\cite{10huang2019mask} adds an additional mask head branch to Mask R-CNN to learn MaskIoU consistent Mask score. The combination of Mask R-CNN and MaskIoU Head solves the problem of mismatch between the confidence score and localization accuracy of predicted masks. These methods have reached state-of-the-art in the field of instance segmentation.
	
	\subsection{Amodal Instance Segmentation}
	The task of amodal segmentation was first proposed by~\cite{1li2016amodal}. They use the modally annotated data for object overlap data enhancement to generate amodal data and then used it to train and validate their methods. They proposed the first method for amodal segmentation, which expands the bounding box of the instance and regenerates the heat map. With the release of some amodal annotation datasets, the research process of amodal segmentation has been accelerated.~\cite{2zhu2017semantic} uses an amodal annotated dataset to train ShapeMask \cite{14pinheiro2015learning}, gets AmodalMask as the baseline. ORCNN \cite{4follmann2019learning} can directly predict the amodal mask and visible mask of the instance by adding the mask branch of Mask R-CNN \cite{6he2017mask}. The former subtracts the latter to get the occluded part. ASN \cite{3qi2019amodal} adds a branch to determine whether the instance is occluded and performs multi-level encoding of the determination result with the RoI feature map before predicting the amodal mask and then performs amodal segmentation. SLN \cite{5zhang2019learning} introduces a semantic-aware distance map instead of the mask as the prediction target to segment the amodal mask of the instance. VRS\&SP \cite{9yuting2021amodal} proposes to simulate human amodal perception, first roughly estimate the visible mask and amodal mask, and then use the shape prior to refining the amodal mask.
	
	\section{Methods}
	\subsection{The Architecture of ARCNN}
\begin{figure*}
		\begin{center}
			\includegraphics[width=12cm]{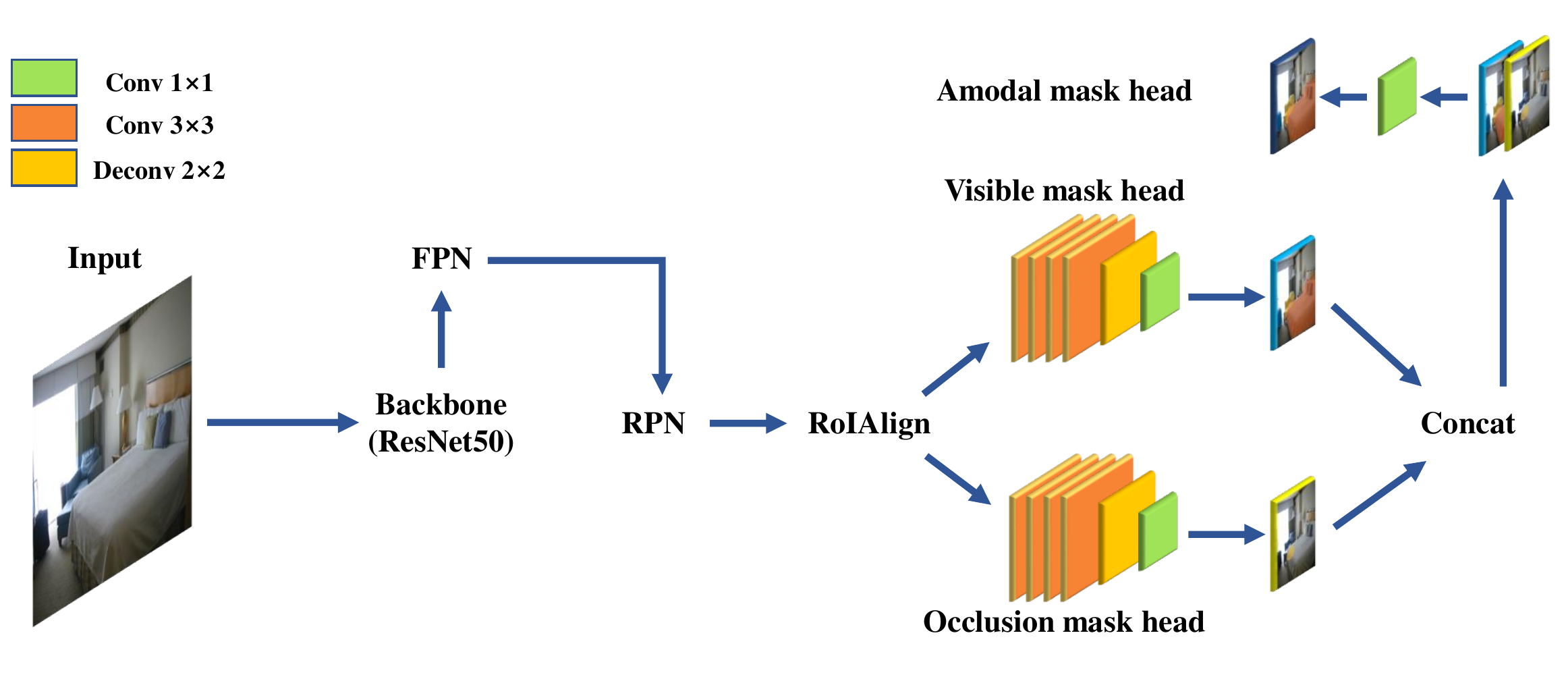}
		\end{center}
		\caption{The architecture of ARCNN}
		\label{fig2}
	\end{figure*}

	We name our method Amodal R-CNN (ARCNN), as this architecture is based on MaskRCNN (MRCNN) \cite{6he2017mask}. Inspired by Occlusion R-CNN (ORCNN) \cite{4follmann2019learning}, we extend MRCNN with two additional heads to predict amodal masks (amodal mask head) and the occlusion masks (occlusion mask head). As for the original mask head of MRCNN, it’s used to predict visible masks(visible mask head). The architecture of ARCNN is shown in Fig. \ref{fig2}.
	
	Firstly, image features are extracted by the backbone (ResNet50 \cite{23he2016deep})). After these features are sent to FPN \cite{13lin2017feature} for multi-scale fusion, they are input into the RPN to generate proposals. The proposals then are sent to RoIAlign to get RoIs as the input of visible mask head and occlusion mask head. Finally, the output of visible mask head and occlusion mask head are concated, then sent to the amodal mask head to obtain the amodal mask. Among them, in order to make the generated proposals can include the visible mask and occlusion mask of the instance, the RPN is trained with the bounding box of the amodal instance as the ground truth.
	
	The visible mask head and occlusion mask head have the same structure, that is, four cascaded 3*3 convolutional layers, a 2*2 deconvolutional layer with stride 2, a 1*1 convolutional layer. These convolutional layers are used to predict masks by using features from RoIs. The amodal mask head is a 1*1 convolutional layer.
	
	We propose the method to concat the visible mask and occlusion mask of the instance in a jigsaw-like operation so that we can decompose the prediction of the amodal mask into the prediction of the visible mask and the occlusion mask and make each branch focus on the modeling of the object. So as to better cope with the challenges brought by the complex task of amodal segmentation.
	
	\subsection{Modeling of Occlusion Relationship}
	The reason for the occlusion in the image is the overlap of two objects. And in the real world, this kind of overlap often contains certain rules. For example, in the occlusion relationship formed by the dinner plate and bread, it is often that the bread obscures the dinner plate. This is a kind of occlusion context information. Therefore, we use a 1*1 convolutional layer (amodal mask head) to model the relationship between the masks, thereby improving the model's ability to obtain occlusion context information during the amodal segmentation process. The modeling process is shown in Fig. \ref{fig3}. The modeling relationship is as follows:
	\begin{figure*}
		\begin{center}
			\includegraphics[width=12cm]{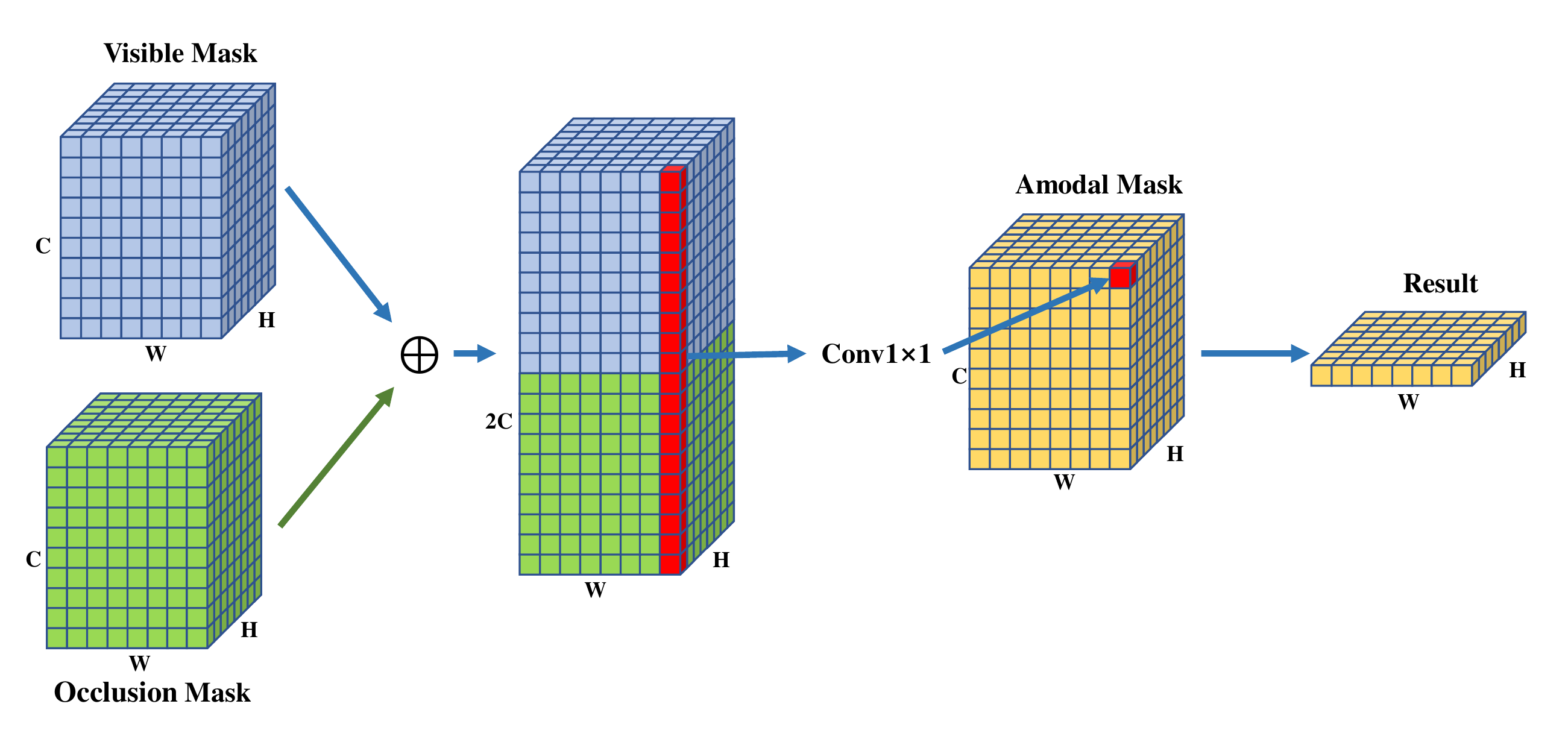}
		\end{center}
		\caption{Visible Mask and Occlusion Mask are the output from the visible and occluded mask branches, respectively. The number of channels of the tensor C is the number of categories of instances in the dataset, and W and H are tensors, respectively, Width and height. Amodal Mask is the mask predicted by the category of all instances in the dataset. Finally, according to the result of the classification, the mask of the corresponding class is selected as the output result.}
		\label{fig3}
	\end{figure*}
	
	\begin{equation}
	A{M_T} = \sum\limits_{i = 1}^n {V{W_{{T_i}}} \cdot V{M_i} + O{W_{{T_i}}} \cdot O{M_i}}
	\end{equation}
	
	Among them, $A{M_T}$ is the corresponding T-th type mask in the C*H*W tensor of amodal mask. The n is the number of instance categories owned by the dataset. $V{M_i}$, $O{M_i}$ are respectively the visible mask and the occluded mask of the i-th type output by the mask branch. $V{W_{{T_i}}}$, $O{W_{{T_i}}}$ are the weights learned by the 1*1 convolutional layer that represent the relationship between the visible and occluded masks of the T-th object and the i-th object.
	
	Due to the modeling of the occlusion relationship, the model can make full use of the occlusion context information.
	
	\subsection{Loss Function}
	The final prediction output of our model includes the bounding box, category, amodal mask, visible mask, and occlusion mask of the instance. These five parts are interrelated, and any part will affect the accuracy of the model. In order to coordinate the model as a whole, we assign the same weight to these five parts of loss. 
	
	The final loss function L:
	\begin{equation}
	L{\rm{ }} = {\rm{  }}{L_{box}}{\rm{ +  }}{L_{cls}}{\rm{  +  }}{L_{AM}}{\rm{  +  }}{L_{VM}}{\rm{  +  }}{L_{OM}}
	\end{equation}
	
	Among them, $ L_{box} $ adopts Smooth L1 loss; $ L_{cls} $  adopts cross entropy loss; $ L_{AM} $, $ L_{VM} $, and $ L_{OM} $ all adopt binary cross entropy loss.
	
	\section{Experiments} 
	\subsection{Datasets}
	Our experiments are conducted on the following three amodal annotated datasets: the KINS dataset \cite{3qi2019amodal} and the COCOA cls dataset \cite{4follmann2019learning}.
	
	The KINS dataset is based on the KITTI dataset \cite{8geiger2012we} for autonomous driving. It consists of 7474 images in the training set and 7517 images in the validation set. The KINS dataset has seven categories of instances.
	The COCOA cls dataset is based on the COCOA dataset \cite{2zhu2017semantic} and COCO dataset \cite{12lin2014microsoft} about the complex everyday scenes. It consists of 2476 training images and 1223 validation images. This dataset has 80 categories of instances.
	\subsection{Experimental Details}
	We use detectron2 to build our model. All experiments are done on a GPU with a model of GeForce GTX 1080Ti and a memory of 11G.
	
	The main hyperparameters are set as follows: For the KINS dataset, batch size: 4, learning rate: 0.0025, iteration: 48000. For the COCOA cls dataset, batch size: 2, learning rate: 0.0005, iteration: 10000. Model training adopts the Stochastic Gradient Descent \cite{22zinkevich2010parallelized} strategy. The backbone of the model in the experiment is resnet50 \cite{23he2016deep}.
	
	\subsection{Evaluation Criterion}
	In order to make the evaluation of the model in the amodal segmentation task have universal significance, we choose the average precision (AP) and average recall (AR) as metrics that are commonly used in the instance segmentation task. Among them, due to most of the occlusion mask has a small area, the deviation of a few pixels may make a huge difference with the ground truth IoU. Therefore, we calculate the AP of the amodal mask of instances where the occlusion rate exceeds 15$\%$ to reflect the model’s ability to predict the occlusion. For fair comparisons, We use the evaluation API of the COCO dataset \cite{12lin2014microsoft}.
	
	\subsection{Baselines}
	\begin{itemize}
		\item \textbf{ORCNN} \cite{4follmann2019learning} adds a branch to the Mask R-CNN, and the two branches respectively predict amodal mask and visible mask. Subtract the visible mask from the amodal mask to obtain the occlusion mask, thereby completing the task of amodal segmentation.
		\item \textbf{VRS\&SP} \cite{9yuting2021amodal} firstly estimates a coarse visible mask and a coarse amodal mask. Then based on the coarse prediction, it infers the amodal mask by concentrating on the visible region and utilizing the shape prior in the memory.
	\end{itemize}
	
	\subsection{Experimental Results}
	We have completed the experiments of the method we proposed on three datasets and the reproduction of ORCNN \cite{4follmann2019learning}. The experimental results of the VRS\&SP model are quoted from VRS\&SP \cite{9yuting2021amodal}. The performance of these models is shown in Table \ref{table:1}.
	
	\subsubsection{Quantitative Analysis}
	We have carried out the following three comparisons and analyses.
	
	\textbf{ARCNN $ VS $ ORCNN.} It can be seen from the table that the evaluation indicators of the amodal mask and the occluded mask segmented by ARCNN on the two datasets exceed ORCNN. And for the COCOA cls dataset, ARCNN significantly surpasses ORCNN in the performance of amodal mask and occluded mask. For the visible mask prediction, there is only a small difference (less than 0.3) between the two indicators. This shows that our proposed method surpasses ORCNN in the performance of amodal segmentation.
	
	\textbf{ARCNN-add $ VS $ ORCNN.} ARCNN-add is a method of directly adding the output of the visible and occluded branches to get the amodal mask. It has a similar network composition to ORCNN. But it has roughly the same performance as ORCNN on the KINS dataset. On the COCO cls dataset, ARCNN-add's indicators fully exceed ORCNN. This shows that our jigsaw-like idea is effective in improving performance on amodal segmentation tasks.
	
	\textbf{ARCNN $ VS $ VRS\&SP.} VRS\&SP is a state-of-the-art method that introduces shape priors. It can be seen from the experimental results that, except for the occluded mask evaluation indicators of the COCOA cls dataset, the ARCNN we proposed exceeds VRS\&SP in all indicators. This shows that our proposed method exceeds the current state-of-the-art methods.
	
	\subsubsection{Ablation Studies}
	We designed the method ARCNN-add for ablation experiments. The difference between ARCNN and ARCNN-add is that ARCNN not only stitches visible and invisible masks based on a jigsaw-like idea    but also models the occlusion relationship between category instances. In terms of indicators, ARCNN surpasses ARCNN-add in both amodal mask and occluded mask. In terms of visible mask-related evaluation indicators, the gap between the two methods is very small. This shows that our proposed occlusion relationship modeling, using the context information of occlusion, can improve the performance of the model.
	\begin{table}[]
		\begin{center}
			\begin{tabular}{cccccc}
				\hline
				\multicolumn{6}{c}{\textbf{KINS}}                                                                                                                                                                                \\ \hline
				\multicolumn{1}{c|}{}                   & \multicolumn{2}{c|}{Amodal}                                               & \multicolumn{2}{c|}{Visible}                                              & Occluded       \\ \hline
				\multicolumn{1}{c|}{}                   & \multicolumn{1}{c|}{AP}             & \multicolumn{1}{c|}{AR}             & \multicolumn{1}{c|}{AP}             & \multicolumn{1}{c|}{AR}             & AP             \\ \hline
				\multicolumn{1}{c|}{ORCNN}   & \multicolumn{1}{c|}{30.57}          & \multicolumn{1}{c|}{19.88}          & \multicolumn{1}{c|}{\textbf{30.95}} & \multicolumn{1}{c|}{20.6}           & 36.15          \\
				\multicolumn{1}{c|}{VRS\&SP} & \multicolumn{1}{c|}{32.08}          & \multicolumn{1}{c|}{20.9}           & \multicolumn{1}{c|}{29.88}          & \multicolumn{1}{c|}{19.88}          & 37.4           \\
				\multicolumn{1}{c|}{ARCNN-add (ours)}   & \multicolumn{1}{c|}{30.2}           & \multicolumn{1}{c|}{19.75}          & \multicolumn{1}{c|}{30.89}          & \multicolumn{1}{c|}{\textbf{20.61}} & 36.27          \\
				\multicolumn{1}{c|}{ARCNN (ours)}       & \multicolumn{1}{c|}{\textbf{32.94}} & \multicolumn{1}{c|}{\textbf{20.96}} & \multicolumn{1}{c|}{30.68}          & \multicolumn{1}{c|}{20.56}          & \textbf{38.71} \\ \hline
				\multicolumn{6}{c}{\textbf{COCOA cls}}                                                                                                                                                                           \\ \hline
				\multicolumn{1}{c|}{}                   & \multicolumn{2}{c|}{Amodal}                                               & \multicolumn{2}{c|}{Visible}                                              & Occluded       \\ \hline
				\multicolumn{1}{c|}{}                   & \multicolumn{1}{c|}{AP}             & \multicolumn{1}{c|}{AR}             & \multicolumn{1}{c|}{AP}             & \multicolumn{1}{c|}{AR}             & AP             \\ \hline
				\multicolumn{1}{c|}{ORCNN}   & \multicolumn{1}{c|}{30.75}          & \multicolumn{1}{c|}{32.55}          & \multicolumn{1}{c|}{34.8}           & \multicolumn{1}{c|}{36.78}          & 18.9           \\
				\multicolumn{1}{c|}{VRS\&SP} & \multicolumn{1}{c|}{35.41}          & \multicolumn{1}{c|}{37.11}          & \multicolumn{1}{c|}{34.58}          & \multicolumn{1}{c|}{36.42}          & \textbf{22.17} \\
				\multicolumn{1}{c|}{ARCNN-add (ours)}   & \multicolumn{1}{c|}{32.26}          & \multicolumn{1}{c|}{34.06}          & \multicolumn{1}{c|}{35.46}          & \multicolumn{1}{c|}{\textbf{37.25}} & 19.32          \\
				\multicolumn{1}{c|}{ARCNN (ours)}       & \multicolumn{1}{c|}{\textbf{36.29}} & \multicolumn{1}{c|}{\textbf{37.39}} & \multicolumn{1}{c|}{\textbf{35.48}} & \multicolumn{1}{c|}{36.69}          & 20.84          \\ \hline
			\end{tabular}
		\end{center}
		\caption{Occluded AP infers to amodal mask AP of the instances whose occlusion rate is more than 15$\%$. ARCNN-add is the method directly adding the visible and occluded output of the branch.}
		\label{table:1}
	\end{table}

	\begin{figure*}
		\begin{center}
			\includegraphics[width=12cm]{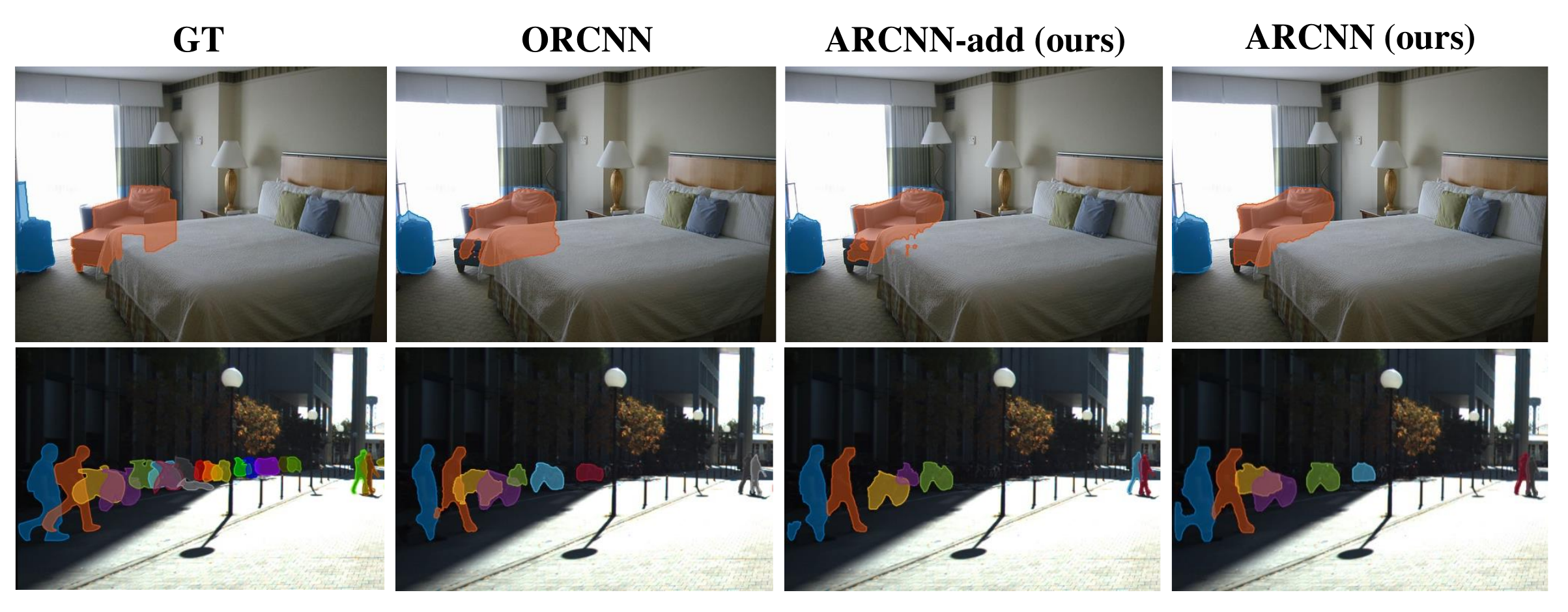}
		\end{center}
		\caption{The columns from left to right are the ground-truth amodal masks, prediction of ORCNN and Ours, respectively. The first row is the result from COCOA cls dataset. And the other row is from KINS dataset.}
		\label{fig5}
	\end{figure*}

	\begin{figure*}[htb]
	\begin{center}
		\includegraphics[width=13cm]{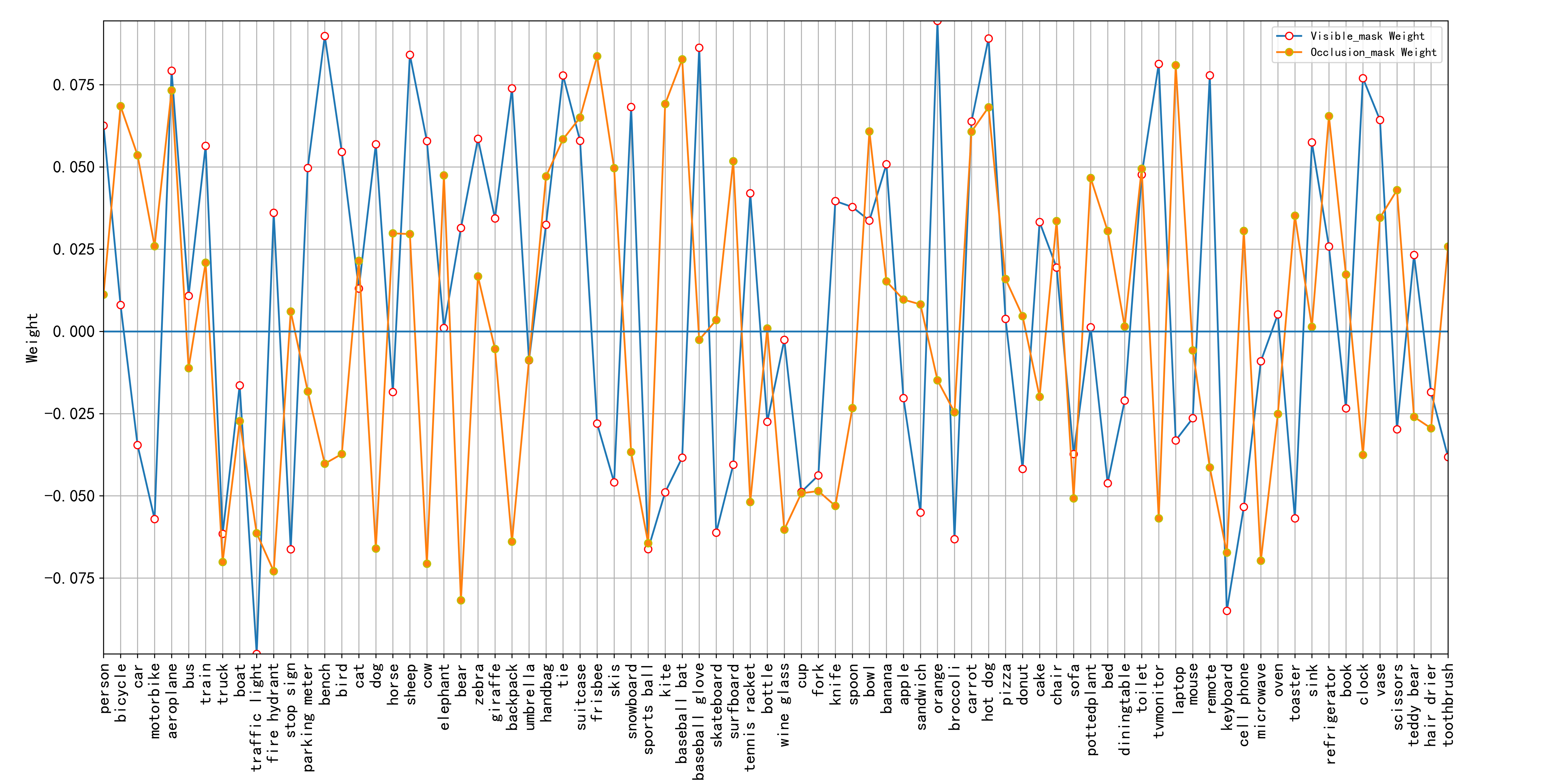}
	\end{center}
	\caption{This is the plot of the cat's 1*1 conv weight. The abscissa represents the category of the instance in the dataset. The ordinate is the weight value.}
	\label{fig4}
	\end{figure*}

	\subsubsection{Visualization of Amodal Results}

	Also, we visualized the amodal results of our method and ORCNN, and the results are shown in Fig. \ref{fig5}. From the comparison of the pictures, we can see that our proposed method is more complete than the amodal mask predicted by ORCNN. And the ARCNN predicts that the amodal mask is smoother than the ARCNN-add. This also proves the effectiveness of our proposed method from another angle.

	\subsection{Visualization of Occlusion Relationship}

	We visualize the weights of the 1*1 convolutional layer between the modeling cat of the model trained under the COCOA cls dataset and other classes as Fig. \ref{fig4}.
	
	The figure contains two types of information. The first category is the correlation between other categories of masks and cat category masks. If the weight is greater than zero, the two are positively correlated, if the weight is less than zero, the two are negatively correlated, and if the weight is equal to zero, the two are uncorrelated. The second category is the relationship between masks of other categories and the occlusion order of the cat category of masks. $V{W_{{T_i}}}$, $O{W_{{T_i}}}$ respectively represent the possibility of the cat being occluded by the i-th category and cat occluding the i-th category. The relative size between the two reflects the relationship between other categories and the occlusion order of cat on the entire dataset to a certain extent. In the figure, the laptop corresponds to $V{W_{{T_i}}}$ < 0, $O{W_{{T_i}}}$ > 0, that is, the occlusion order that appears on the entire dataset is cat occluding the laptop. This is also confirmed in the process of visualizing the picture of the dataset. In the figure, the person corresponding to $V{W_{{T_i}}}$ > $O{W_{{T_i}}}$ > 0, that is, the relationship between person occluding cat and cat occluding person has appeared in the entire dataset, but the former appears more often.
	
	\section{Conclusions}
	In this paper, we propose a method of decomposing the task of amodal segmentation into the visible mask and occlusion mask prediction, and finally stitching the two parts to obtain the amodal mask. The predictions of these two parts are naturally decoupled. In this way, the division of labor of the network branches can be clearly realized so as to make each branch focus on the modeling of the object. And we believe that there are certain rules in the occlusion relationship in the real world, so we modeled it and applied the modeling results to obtain the amodal mask. Experimental results prove that our proposed method is simple and effective. The performance of our proposed method on amodal segmentation tasks exceeds the existing state-of-the-art methods.
	
	\bibliography{egbib}
\end{document}